\documentclass[twocolumn]{IEEEtran}

\def\BibTeX{{\rm B\kern-.05em{\sc i\kern-.025em b}\kern-.08em
    T\kern-.1667em\lower.7ex\hbox{E}\kern-.125emX}}
\usepackage[cmex10]{amsmath}
\usepackage{amsfonts}
\usepackage[dvips]{graphicx}

\begin{document}

\title{Higher Order Moments Generation by Mellin Transform for Compound Models of
  Clutter}
\author{C. Bhattacharya 
\\email: {cbhat0@ieee.org}
\\ \textrm{DEAL(DRDO), Dehradun 248001, India }}

 \maketitle
 \begin{abstract}
 The compound models of clutter statistics are found
 suitable to describe the nonstationary nature of radar backscattering
 from high-resolution observations. In this letter, we show that the
 properties of Mellin transform can be utilized to generate higher order
 moments of simple and compound models of clutter statistics in a compact manner.
\end{abstract}
\section{Introduction}
\PARstart{R}{adar} backscattering from ground or sea surfaces are
wide-sense stationary for low-resolution observations as
expectations of clutter statistics or moments are assumed to be
independent of spatio-temporal changes. For high-resolution
observations, such surfaces reveal heterogeneous structures such
as swell in sea waves or winds blowing over the canopy of
grasslands that result in nonstationary clutter statistics [1],
[2], [4]. The compound models of probability density functions
(pdf) incorporate the variation in the parameters of clutter in
such cases.

Traditionally higher order moments of a continuous random variable
(rv) \textbf{\emph{X}} are generated from higher order derivatives
of its characteristic function defined as

\begin{equation}
 \label{one}
  \Phi_\emph{\textbf{X}}(\omega)=\emph{E}\{\exp(\emph{j}\omega\emph{x})\}
              =\int_{-\infty}^{\infty}\emph{f}_{\textbf{\emph{X}}}(\emph{x})
                \exp(\emph{j}\omega\emph{x})\emph{dx}\textrm{.}
\end{equation}
\\ The continuous pdf $\emph{f}_{\textbf{\emph{X}}}(\emph{x})$
 is for $-\infty<\emph{x}<\infty$. Generation of moments and cumulants
from (\ref{one}) for the compound models of clutter require
solutions of incomplete integrals. The domain of \textbf{\emph{X}}
is $0\leq\emph{x}<\infty$ for amplitude and power statistics, and
$\int_{0}^{\infty}\emph{f}_{\textbf{\emph{X}}}(\emph{x})\emph{dx}=1$.
Properties of Mellin transform  provide the formalism to derive
higher order moments in a compact manner in such cases. Some of
these properties were used in [3], [6] to derive the moments for
high-resolution synthetic aperture radar (SAR) clutter statistics.
Here, we show that the properties of Mellin transform can be
utilized in an effective manner for both simple and compound
models of clutter either in amplitude or in intensity domain.
\section{Mellin Transform Properties}
Mellin transform exists for a continuous function
$\emph{f}_{\textbf{\emph{X}}}(\emph{x})$ defined over
$\mathbb{R}_{+}$. The transform operator is the second kind
characteristic function $\Phi_\emph{\textbf{X}}(s)$ expressed as

\begin{equation}
 \label{two}
\Phi_{\emph{\textbf{X}}}(s)=\mathcal{M}[\emph{f}_{\textbf{\emph{X}}}(\emph{x});\emph{s}]
 =\int_{0}^{\infty}\emph{x}^{\emph{s}-1}\emph{f}_{\textbf{\emph{X}}}(\emph{x})\emph{dx}
\textrm{.}
\end{equation}
Here $\emph{s}=\emph{a}+\emph{j}\emph{b}\in\mathbb{C}$ is the
complex Laplace transform variable. Traditional moments are
generated from (\ref{two}) with
$\emph{s}=\emph{n}+1,\emph{n}\in\mathbb{Z}_{+}.$

\begin{equation}
 \label{three}
\emph{m}_{\emph{n}}=\mathcal{M}{\displaystyle\biggl[\emph{f}_{\textbf{\emph{X}}}(\emph{x});\emph{s}\biggr]}_
{\emph{s}=\emph{n}+1}\textrm{.}
\end{equation}
\\Second-kind moments or the \textit{log-moments} are generated for logarithm
of rv \textbf{\emph{X}} by using the derivative property of Mellin
transform.

\setlength{\arraycolsep}{0.0em}
\begin{eqnarray}
 \label{four}
\tilde{\emph{m}_{\emph{n}}}=\mathcal{M}[\log(\emph{x})^\emph{n}\emph{f}_{\textbf{\emph{X}}}(\emph{x});\emph{s}]
\bigg|_{\emph{s}=1}
=\int_{0}^{\infty}\emph{x}^{\emph{s}-1}\log(\emph{x})^\emph{n}\emph{f}_{\textbf{\emph{X}}}(\emph{x})\emph{dx}
\nonumber\\
\qquad =\frac{d^{\emph{n}}}{d\emph{s}^{\emph{n}}}
\Phi_\emph{\textbf{X}}(\emph{s})\bigg|_{\emph{s}=1}\textrm{.}\qquad\qquad
 \end{eqnarray}

 Analogous to the cumulants derived from logarithm of characteristic function in (\ref{one}),
 the \textit{n}-th order cumulants of second kind or the \textit{log}-\textit{cumulants} are obtained
 from  derivatives of logarithm of $\Phi_\emph{\textbf{X}}(s)$;  i.e., $\Psi_\emph{\textbf{X}}(s)= \log(\Phi_\emph{\textbf{X}}(s))$.

\begin{equation}
 \label{five}
\tilde{\emph{k}_{\emph{n}}}=\frac{d^{\emph{n}}}{d\emph{s}^{\emph{n}}}
 \Psi_\emph{\textbf{X}}(\emph{s})\bigg|_{\emph{s}=1}\textrm{.}
\end{equation}
\\The log-moments and the log-cumulants are related as

\setlength{\arraycolsep}{0.0em}
 \begin{eqnarray}
 \label{six}
 &&\tilde{\emph{k}_{1}}= \tilde{\emph{m}_{1}}\nonumber\\
 &&\tilde{\emph{k}_{2}}= \tilde{\emph{m}_{2}}-\tilde{\emph{m}_{1}}^{2}\nonumber\\
&& \tilde{\emph{k}_{3}}=
\tilde{\emph{m}_{3}}-3\tilde{\emph{m}_{1}}\tilde{\emph{m}_{2}}+2\tilde{\emph{m}_{1}}^{3}
\nonumber\\
&&
\tilde{\emph{k}_{4}}=\tilde{\emph{m}}_{4}-4\tilde{\emph{m}_{1}}\tilde{\emph{m}_{3}}
+6\tilde{\emph{m}_{1}}^{2}\tilde{\emph{m}_{2}}-
3\tilde{\emph{m}_{1}}^{4}\textrm{.}
\end{eqnarray}

 The underlying mean of speckle component of clutter vary
widely in the compound models of amplitude or power statistics
resulting in long-tailed distributions. Speckle arises from
randomness in the distribution of backscattering elements in the
resolution cell, the number of such scatterers is nonstationary
for high-resolution observations. The pdf of high-resolution
clutter is described by taking into account of a rv
\emph{\textbf{Z}} signifying randomness in the mean of clutter.

\begin{equation}
 \label{seven}
\qquad \emph{f}_{\textbf{\emph{X}}}(\emph{x})=
\int_{0}^\infty\emph{f}_\textbf{\emph{X}}(\emph{x}|\emph{z})\emph{f}_\textbf{\emph{Z}}(\emph{z})\emph{dz}
 {}{}, \qquad \emph{z}>0\textrm{.}
\end{equation}
\\The compound pdf model in (\ref{seven}) is a Mellin convolution. One nice property of Mellin transform is the product
form of the components of pdf in the transform domain [5].

\begin{equation}
 \label{eight}
\mathcal{M}[\emph{f}_{\textbf{\emph{X}}}(\emph{x});\emph{s}]=
 \mathcal{M}[\emph{f}_\textbf{\emph{X}}(\emph{x}|\emph{z});\emph{s}]\mathcal{M}[\emph{f}_\textbf{\emph{Z}}(\emph{z});\emph{s}]
\textrm{.}
\end{equation}
\\The log-cumulants of the components in (\ref{eight}) are therefore additive.

\begin{equation}
 \label{nine}
\tilde{\emph{k}_{\emph{n},\emph{x}}}=
\tilde{\emph{k}_{\emph{n},(\emph{x},\emph{z})}}  +
\tilde{\emph{k}_{\emph{n},\emph{z}}}\textrm{.}
\end{equation}
\section{Moments Generation for Simple Models of Clutter}
The shape and scale parameters of simple models of pdf for
low-resolution cases are stationary. The usual pdf of speckle
power is a gamma distribution resulting from convolution of
\emph{L} independent exponential distributions.

\begin{equation}
 \label{ten}
\emph{f}_{\textbf{\emph{V}}}(\emph{v})=\frac{1}{\Gamma(\emph{L})}{\bigg(\frac{\emph{L}}{\mu}\bigg)}^{\emph{L}}
\emph{v}^{(\emph{L}-1)}\exp\bigg(-\frac{\emph{L}\emph{v}}{\mu}\bigg),\qquad\emph{v}\geq
0 \textrm{.}
\end{equation}
\\Here $\Gamma(.)$ is the standard gamma function. The shape and
scale of distribution are determined by $\emph{L}$ and $\mu$, mean
value of clutter power respectively. Corresponding amplitude
distribution turns out to be a Nakagami pdf [2], [6].

\begin{equation}
 \label{eleven}
\emph{f}_{\textbf{\emph{N}}}(\emph{r})=\frac{2}{\Gamma(\emph{L})}\bigg(\frac{{\sqrt\emph{L}}}{\mu}\bigg)^{2\emph{L}}
\emph{r}^{(2\emph{L}-1)}\exp\bigg(-\frac{\emph{L}\emph{r}^{2}}{\mu^{2}}\bigg),\quad\emph{r}\geq
0\textrm{.}
\end{equation}
\\Mellin transform for gamma pdf is

\begin{equation}
 \label{twelve}
\Phi_\emph{\textbf{G}}(\emph{s})=\frac{\lambda^{\emph{L}}}{\Gamma(\emph{L})}
\int_{0}^{\infty}\emph{v}^{(\emph{L+s}-1)-1}\exp(-\lambda\emph{v})\emph{dv}
\end{equation}
\\with $\lambda=\frac{\emph{L}}{\mu}$. Using the transform pair

\begin{displaymath}
\mathcal{M}[\emph{x}^{\emph{u}}\exp(-\lambda\emph{x});\emph{s}]
 \Longleftrightarrow \lambda^{-(\emph{s}+\emph{u})}\Gamma(\emph{s}+\emph{u})
\end{displaymath}
\\we obtain,

\begin{equation}
 \label{thirteen}
 \Phi_\emph{\textbf{G}}(\emph{s})=\bigg(\frac{\mu}{\emph{L}}\bigg)^{\emph{s}-1}
\frac{\Gamma(\emph{s}+\emph{L}-1)}{\Gamma(\emph{L})}\textrm{.}
\end{equation}
\\The moments of first kind for gamma pdf are generated from (\ref{thirteen}) with $\emph{s}=\emph{n}+1$ as

\begin{equation}
 \label{fourteen}
\emph{m}_{\emph{n}}=\Phi_\emph{\textbf{G}}(\emph{s})\bigg|_{\emph{s}=\emph{n}+1}=
\bigg(\frac{\mu}{\emph{L}}\bigg)^{\emph{n}}\frac{\Gamma(\emph{L}+\emph{n})}{\Gamma(\emph{L})}
\textrm{.}
\end{equation}
\\As a special case of the result in (\ref{fourteen}), the  moments of exponential
pdf (for $\emph{L}=1$) are $\emph{m}_{\emph{n}}=\mu^{\emph{n}}
{\emph{n}}!$. Maxwell pdf is the case for $\emph{L} = 3$.

\begin{equation}
 \label{fifteen}
\emph{f}_{\textbf{\emph{M}}}(\emph{u})=\frac{1}{\sigma^{3}}\sqrt{\frac{2}{\pi}}\emph{u}^{2}
\exp\bigg(-\frac{\emph{u}^{2}}{2\sigma^{2}}\bigg),\qquad
\emph{u}\geq 0\textrm{.}
\end{equation}
\\We use the additional Mellin transform pair

\begin{displaymath}
\mathcal{M}[\exp(-\lambda\emph{x}^{2});\emph{s}]
\Longleftrightarrow
\frac{1}{2}(\lambda)^{-\frac{s}{2}}\Gamma(\frac{\emph{s}}{2})
\end{displaymath}
with $\lambda = \frac{1}{2\sigma^{2}}$ ; so that

\begin{equation}
 \label{sixteen}
\Phi_\emph{\textbf{M}}(\emph{s})=\frac{1}{\sigma}\sqrt{\frac{2}{\pi}}(2\sigma^{2})^{\frac{\emph{s}}{2}}
{\frac{\emph{s}}{2}}\Gamma\bigg(\frac{\emph{s}}{2}\bigg)\textrm{.}
\end{equation}
\\The moments of first kind for Maxwell pdf are

\begin{equation}
 \label{seventeen}
\emph{m}_{\emph{n}}=\Phi_\emph{\textbf{M}}(\emph{s})\bigg|_{\emph{s}=\emph{n}+1}=
\frac{1}{\sigma}\sqrt{\frac{2}{\pi}}(2\sigma^{2})^{\frac{\emph{n}+1}{2}}\bigg(\frac{\emph{n}+1}{2}\bigg)
\Gamma\bigg(\frac{\emph{n}+1}{2}\bigg)\textrm{.}
\end{equation}

The moments for amplitude distributions are also derived by Mellin
transform. As for Nakagami distribution, with
$\lambda=\frac{\sqrt{\emph{L}}}{\mu}$

 \setlength{\arraycolsep}{0.0em}
\begin{eqnarray}
 \label{eighteen}
\Phi_\emph{\textbf{N}}(\emph{s})=\frac{2}{\Gamma(\emph{L})}(\lambda^{2})^{\emph{L}}
\int_{0}^{\infty}\emph{r}^{(\emph{s}+2\emph{L}-1)-1}\exp(-\lambda^{2}\emph{r}^{2})\emph{dr}
\nonumber\\
\qquad\qquad
=\bigg(\frac{\sqrt{\emph{L}}}{\mu}\bigg)^{-(\emph{s}-1)}\frac{\Gamma(\emph{L}+\frac{\emph{s}-1}{2})}{\Gamma(\emph{L})}\textrm{.}
\qquad\qquad\qquad
\end{eqnarray}
\\The log-cumulants are easier to derive here. In general the log-cumulants of
Nakagami distribution are derived from (\ref{five})

\begin{equation}
 \label{nineteen}
\tilde{\emph{k}_{\emph{n}}}=\bigg(\frac{1}{2}\bigg)^{\emph{n}}\mathbf{\Upsilon}(\emph{n}-1,\emph{L})
\textrm{.}
\end{equation}
\\Here $\mathbf{\Upsilon}(.)$ is the Digamma function; i. e., the first derivative
of  $\ln\Gamma(\emph{s})$ at $\emph{s} =1$. In general
$\mathbf{\Upsilon}(\emph{n}-1,\emph{L})$ is the $\emph{n}$th
derivative of the Digamma function for variable $\emph{L}$.

One long-tailed pdf often used in sea-clutter amplitude modelling
[1] is Weibull distribution.

\begin{equation}
 \label{twenty}
\emph{f}_{\textbf{\emph{W}}}(\emph{x};\emph{z},\emph{b})=\bigg(\frac{\emph{b}}{\emph{z}}\bigg)
\bigg(\frac{\emph{x}}{\emph{z}}\bigg)^{\emph{b}-1}\exp
{\displaystyle\bigl[-\bigg(\frac{\emph{x}}{\emph{z}}\bigg)^{b}\bigr]},{}{}
\emph{x}\geq0;\emph{b},\emph{z}>0\textrm{.}
\end{equation}
\\Here $\emph{z}$ is the scale parameter and $\emph{b}$ is the
shape parameter of distribution. Mellin transform of
(\ref{twenty}) is

\begin{equation}
 \label{twentyone}
\Phi_\emph{\textbf{W}}(\emph{s})=\frac{\emph{b}}{\emph{z}^{\emph{b}}}\int_{0}^{\infty}
\emph{x}^{(\emph{s}+\emph{b}-1)-1}\exp
{\displaystyle\bigl[-\bigg(\frac{\emph{x}}{\emph{z}}\bigg)^{b}\bigr]}\emph{dx}\textrm{.}
\end{equation}
\\From the Mellin transform pair
\begin{displaymath}
 \mathcal{M}[{\exp(-\lambda\emph{x}^{\emph{b}})};\emph{s}]
 \Longleftrightarrow \emph{b}^{-1}\lambda^{-\frac{\emph{s}}{\emph{b}}}\Gamma(\frac{\emph{s}}{\emph{b}}) ,
\end{displaymath}
  the second characteristic function is

\begin{equation}
 \label{twentytwo}
\Phi_\emph{\textbf{W}}(\emph{s})=\emph{z}^{(\emph{s}-1)}\Gamma
\big(\frac{\emph{s}+\emph{b}-1}{\emph{b}}\big)\textrm{.}
\end{equation}
\\The moments of first kind for Weibull distribution are

\begin{equation}
 \label{twentythree}
\emph{m}_{\emph{n}}=\Phi_\emph{\textbf{W}}(\emph{s})\bigg|_{\emph{s}=
\emph{n}+1}=\emph{z}^{\emph{n}}\Gamma\big(\frac{\emph{n}+\emph{b}
}{\emph{b}}\big)\textrm{.}
\end{equation}
\\The common Rayleigh amplitude pdf is a special case of Weibull
distribution with $b =2$.

\begin{equation}
 \label{twentyfour}
\emph{f}_{\textbf{\emph{R}}}(\emph{r};\emph{z})=
2\bigg(\frac{\emph{r}}{\emph{z}^{2}}\bigg)
\exp{\displaystyle\bigl[-\bigg(\frac{\emph{r}}{\emph{z}}\bigg)^{2}\bigr]},
\qquad \emph{r}\geq0\textrm{.}
\end{equation}
\\The moments of first kind for Rayleigh pdf are

\begin{equation}
 \label{twentyfive}
\emph{m}_{\emph{n}}=\Phi_\emph{\textbf{R}}(\emph{s})
\bigg|_{\emph{s}=\emph{n}+1}=
\emph{z}^{\emph{n}}\Gamma\bigg(\frac{\emph{n}+2}{2}\bigg)\textrm{.}
\end{equation}
\\

We show in the next section the utility of Mellin transform for
deriving the log-moments and the log-cumulants of compound models
of clutter in a compact manner.
\section{Moments Generation for Compound Models of Clutter}
The pdf for compound models of high-resolution clutter have got
two components; pdf of speckle component, and  pdf of the
modulation in mean amplitude or power of speckle. Considering both
to be gamma distributed rv the pdf for generalized gamma
(G$\Gamma$) model of clutter power is [6]

\setlength{\arraycolsep}{0.0em}
\begin{eqnarray}
 \label{twentysix}
\emph{f}_{\textbf{\emph{V}}}(\emph{v})=\frac{1}{\Gamma(\emph{L})\Gamma(\emph{M})}
\bigg(\frac{2\emph{L}\emph{M}}{<\emph{z}>}\bigg)\bigg(\frac{2\emph{L}\emph{M}}{<\emph{z}>}\emph{v}\bigg)
^{(\frac{\emph{L}+\emph{M}-2}{2})}\nonumber\\
\qquad\qquad\qquad
\emph{K}_{{\emph{M}}-{\emph{L}}}{\displaystyle\bigl[2\bigg(\frac{\emph{LM}}{<\emph{z}>}\emph{v}\bigg)
^{\frac{1}{2}}\bigr]}\textrm{.}
\end{eqnarray}
\\
The shape parameter for gamma pdf
$\emph{f}_\textbf{\emph{Z}}(\emph{z})$ of rv $\textbf{\emph{Z}}$
according to (\ref{seven}) is $\emph{M}$, and
$\emph{K}_{\emph{M}-\emph{L}}(.)$ is the second kind modified
Bessel function of order $(\emph{M}-\emph{L})$. The mean estimate
of $<\emph{z}>=\mu$. Assuming speckle and the modulation in mean
power in the high-resolution cell to be independent of each other,
we have by Mellin convolution property in (\ref{eight})

\setlength{\arraycolsep}{0.0em} \setlength{\arraycolsep}{0.0em}
\begin{eqnarray}
 \label{twentyseven}
\Psi_\emph{\textbf{V}}(s)=(\emph{s}-1)\log\bigg(\frac{\mu}{\emph{LM}}\bigg)+
\log\Gamma(\emph{s}+\emph{L}-1)+ \nonumber\\
\qquad\qquad{}{}\log\Gamma(\emph{s}+\emph{M}-1)
-\log\Gamma(\emph{L})-\log\Gamma(\emph{M})\textrm{.}
\end{eqnarray}
\\
The log-cumulants of G$\Gamma$ model are

\setlength{\arraycolsep}{0.0em}\setlength{\arraycolsep}{0.0em}
\begin{eqnarray}
 \label{twentyeight}
&&{} \tilde{\emph{k}_{1}}= \log\bigg(\frac{\mu}{\emph{LM}}\bigg)+
\mathbf{\Upsilon}(\emph{L})+\mathbf{\Upsilon}(\emph{M})\bigg|_{\emph{s}=1}\nonumber\\
&&{}
\tilde{\emph{k}_{\emph{n}}}=\mathbf{\Upsilon}(\emph{n}-1,\emph{L})
+\mathbf{\Upsilon}(\emph{n}-1,\emph{M})\textrm{.}
\end{eqnarray}
\\

Spikes in high-resolution ground-clutter amplitude at low grazing
angles are often described by the K-distribution model [2], [4].
The compound K-pdf

\begin{equation}
 \label{twentynine}
\emph{f}_{\textbf{\emph{N}}}(\emph{r})=\frac{4\emph{b}^{\frac{(\alpha+1)}{2}}\emph{r}^{\alpha}}
{\Gamma(\alpha)}\emph{K}_{{\alpha}-1}(2\emph{r}\sqrt{\emph{b}})
\end{equation}
is a Mellin convolution of Rayleigh pdf and exponential pdf given
by,

\begin{equation}
 \label{thirty}
\emph{f}_{\textbf{\emph{N}}}(\emph{r})=\frac{4\emph{r}\emph{b}^{\alpha}}{\Gamma(\alpha)}
\int_{0}^{\infty}\frac{\emph{dz}}{\emph{z}}\emph{z}^{\alpha-1}\exp\bigg(-\emph{b}\emph{z}
-\frac{\emph{r}^{2}}{\emph{z}}\bigg)\textrm{.}
\end{equation}
\\
Here $\alpha$ is the shape parameter in the variation of mean of
sea or ground clutter amplitude, and $\emph{b}$ is the scale
parameter for associated speckle amplitude. Following derivation
for Nakagami pdf in (\ref{nineteen}) the second characteristics
function for K-pdf is given by,

\begin{equation}
 \label{thirtyone}
\Phi_\emph{\textbf{N}}(\emph{s})=\emph{b}^{-(\frac{\emph{s}-1}{2})}\mu^{\emph{s}-1}
\Gamma\bigg(\frac{\emph{s}}{2}+\frac{1}{2}\bigg)
\frac{\Gamma(\alpha+\frac{\emph{s}-1}{2})}{\Gamma(\alpha)}\textrm{
.}
\end{equation}
where $<\emph{z}>=\mu^{2}$. The log-cumulants for K-pdf are

 \begin{eqnarray}
\Psi_\emph{\textbf{N}}(s)=-\bigg(\frac{\emph{s}-1}{2}\bigg)\log\emph{b}+(\emph{s}-1)\log\mu
+\log\Gamma\bigg(\frac{\emph{s}}{2}+\frac{1}{2}\bigg)+ \nonumber\\
\qquad
 \log\Gamma\bigg(\alpha+\frac{\emph{s}-1}{2}\bigg)-\log\Gamma(\alpha),\nonumber
\end{eqnarray}
and
\setlength{\arraycolsep}{0.0em}
 \begin{eqnarray}
 \label{thirtytwo}
&&{}\tilde{\emph{k}_{1}}= -\frac{1}{2}\log\emph{b}+ \log\mu
+\frac{1}{2}\mathbf{\Upsilon}(\alpha)+
\frac{1}{2}\mathbf{\Upsilon}(1)\bigg|_{\emph{s}=1}\nonumber\\
&&{}\tilde{\emph{k}_{\emph{n}}}=
\bigg(\frac{1}{2}\bigg)^{\emph{n}}\mathbf{\Upsilon}(\emph{n}-1,\alpha)\textrm{.}
\end{eqnarray}
\\
Here $\mathbf{\Upsilon}(1)= -0.577215$ is the Euler constant [5].
This shows that the log-cumulants of K-distribution are determined
by the higher order log-cumulants of Nakagami distribution in the
mean of high-resolution ground or sea clutter.

A more extended case of compound clutter model is the scene where
variation in the shape of clutter amplitude distribution is given
by generalized Weibull distribution [4].

\begin{equation}
 \label{thirtythree}
\emph{f}_{\textbf{\emph{WN}}}(\emph{r};\emph{b},\emph{c},\alpha)=\frac{2\emph{c}\emph{b}^{\alpha}}{\Gamma(\alpha)}
\emph{r}^{\emph{c}-1}\int_{0}^{\infty}\frac{\emph{dz}}{\emph{z}^{\emph{c}}}\emph{z}^{2\alpha-1}
\exp\bigg[-\bigg(\frac{\emph{r}}{\emph{z}}\bigg)^{\emph{c}}-\emph{b}\emph{z}^{2}\bigg]\textrm{.}
\end{equation}
This is a Mellin convolution where randomness in the mean
amplitude of clutter is described by Nakagami pdf with the shape
parameter being $\alpha$.

\begin{displaymath}
\emph{f}_{\textbf{\emph{N}}}(\emph{z};\emph{b},\alpha)=\frac{2\emph{b}^{\alpha}}{\Gamma(\alpha)}\nonumber
\emph{z}^{2\alpha-1}\exp({-\emph{b}\emph{z}^{2}}),
\end{displaymath}
and the clutter amplitude follows a generalized Weibull
distribution with the shape parameter being $\emph{c}$.

\begin{displaymath}
\emph{f}_{\textbf{\emph{W}}}(\emph{r}|\emph{z};\emph{c})=\frac{\emph{c}}{\emph{z}^{\emph{c}}}
\emph{r}^{\emph{c}-1}\exp\bigg[-\bigg(\frac{\emph{r}}{\emph{z}}\bigg)^{\emph{c}}\bigg]\textrm{.}
\end{displaymath}
\\
Following the transform rule of Mellin convolution,
\setlength{\arraycolsep}{0.0em}
 \begin{eqnarray}
 \label{thirtyfour}
&&{}\Phi_\emph{\textbf{W}\textbf{N}}(\emph{s})=\Phi_\emph{\textbf{W}}(\emph{s})\Phi_\emph{\textbf{N}}(\emph{s})
 \qquad \nonumber\\&&{}
 \qquad =\bigg(\frac{\sigma}{\emph{b}}\bigg)^{\frac{\emph{s}-1}{2}}\Gamma\bigg(\frac{\emph{s}+\emph{c}-1}{c}\bigg)
\frac{\Gamma\big(\alpha+\frac{\emph{s}-1}{2}\big)}{\Gamma
(\alpha)}\textrm{.}
\end{eqnarray}
where $<\emph{z}^{2}>=\sigma$. The log-moments for this
generalized Weibull model of clutter according to (\ref{nine}) are

\setlength{\arraycolsep}{0.0em}\setlength{\arraycolsep}{0.0em}
 \begin{eqnarray}
 \label{thirtyfive}
&&{} \tilde{\emph{k}_{1}}=
\frac{1}{2}\log\bigg(\frac{\sigma}{\emph{b}}\bigg)+
\frac{1}{\emph{c}}\mathbf{\Upsilon}(1)+\frac{1}{2}\mathbf{\Upsilon}(\alpha)\bigg|_{\emph{s}=1}\nonumber\\
&&{}
\tilde{\emph{k}_{\emph{n}}}=\bigg(\frac{1}{2}\bigg)^{\emph{n}}\mathbf{\Upsilon}(\emph{n}-1,\alpha)
\textrm{.}
\end{eqnarray}
\\

Another compound model used to describe high-resolution SAR
clutter is the Fisher distribution [3].

\begin{equation}
 \label{thirtysix}
\emph{f}_{\textbf{\emph{F}}}(\emph{u})=\frac{\Gamma(\emph{L}+\emph{M})}{\Gamma(\emph{L})\Gamma(\emph{M})}
 \bigg(\frac{\emph{L}}{\emph{M}\mu}\bigg)\frac{\big(\frac{\emph{L}}{\emph{M}\mu}\emph{u}\big)^{\emph{L}-1}}
 {\big(1+\frac{\emph{L}}{\emph{M}\mu}\emph{u}\big)^{\emph{L}+\emph{M}}}\textrm{.}
\end{equation}
\\
Consider $\frac{\emph{L}}{\emph{M}\mu}\emph{u}=\lambda$; following
the Mellin transform pair

\begin{displaymath}
 \mathcal{M}[(1+\lambda)^{-\emph{b}};\emph{s}]
 \Longleftrightarrow
 \frac{\Gamma(\emph{s})\Gamma(\emph{b}-\emph{s})}{\Gamma(\emph{b})},
  \textrm{ with } \emph{b}=\emph{L}+\emph{M}
\end{displaymath}
the second characteristic function for Fisher distribution is

\begin{equation}
 \label{thirtyseven}
\Phi_\emph{\textbf{F}}(\emph{s})=
\bigg(\frac{\emph{M}\mu}{\emph{L}}\bigg)^{\emph{s}-1}\frac{1}{\Gamma(\emph{L})(\emph{M})}
\Gamma(\emph{s}+\emph{L}-1)\Gamma(\emph{M}+1-\emph{s})\textrm{.}
\end{equation}
\\Corresponding log-cumulants are

\setlength{\arraycolsep}{0.0em}\setlength{\arraycolsep}{0.0em}
 \begin{eqnarray}
 \label{thirtyeight}
&&{} \tilde{\emph{k}_{1}}= \log\mu
+[\mathbf{\Upsilon}(\emph{L})-\log\emph{L}]+[\mathbf{\Upsilon}(\emph{M})-\log\emph{M}]
\bigg|_{\emph{s}=1}\nonumber\\
&&{}
\tilde{\emph{k}_{\emph{n}}}=\mathbf{\Upsilon}(\emph{n}-1,\emph{L})+(-1)^{\emph{n}}\mathbf{\Upsilon}
(\emph{n}-1,\emph{M})\textrm{.}
\end{eqnarray}

One useful application of the log-cumulants and their relationship
with the log-moments in (\ref{six}) is estimation of parameters of
texture. Empirical data from high-resolution radar backscattering
$\emph{x}(\emph{t})$ follow the product model,

\begin{equation}
 \label{thirtynine}
\emph{x}(\emph{t})=\emph{u}(\emph{t})\emph{z}(\emph{t})\textrm{.}
\end{equation}
Here $\emph{u}(\emph{t})$ is the speckle component, and
$\emph{z}(\emph{t})$ represent texture signifying variation in the
mean parameter. The log-moments of observed data and the
log-cumulants of texture can be estimated for different compound
models utilizing the relationships in (\ref{four}) and
(\ref{six}). Parameters for texture are derived using the
log-cumulants of speckle as in (\ref{nine}), and can be verified
with the theoretical values derived in the paper. For example,
second and fourth order log-cumulants of texture component for
G$\Gamma$ model of high-resolution ground clutter in
(\ref{twentyeight}) are estimated in Fig. 1. Second and fourth
order log-moments of $\emph{x}(\emph{t})$ are derived from the
log-cumulants of $\emph{z}(\emph{t})$ assuming it to be a gamma
variable, and $\emph{u}(\emph{t})$ also follows gamma
distribution. The results of simulation show that higher order
log-cumulants of texture vanish with increasing values of shape
parameter $\emph{M}$. This is expected in the present case as the
texture component follows a nearly Gaussian distribution with
constant mean for increasing values of $\emph{M}$. For values of
$\emph{M}<1$, there is presence of large amount of spikes in
 observed data signifying high values of log-moments and cumulants.
Log-moments of clutter tend to become constant with increasing
$\emph{M}$ signifying stationarity of low-resolution observations.

\section{Conclusion}
The utility of Mellin transform properties to generate higher
order moments of simple and compound models of clutter in both
amplitude and power domain is shown in this letter. The second
kind characteristic function and its properties provide compact
analytical expressions for higher order moments that are useful to
interpret texture properties of high-resolution clutter.
\\
\\
\begin{figure}
\centering
\includegraphics[width=3.5in, height =3.0in]{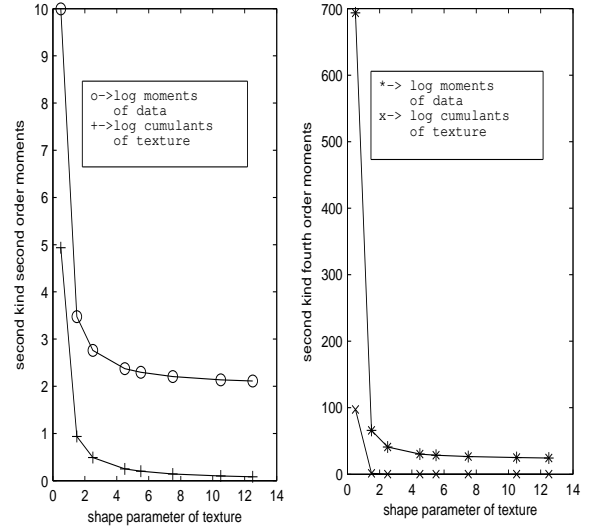}
 \caption{Log-moments of data and log-cumulants of texture for simulation of G$\Gamma$ model
 of high-resolution clutter. Left: second order moments; right: fourth order moments.}
\end{figure}

\section*{Acknowledgment}
\addcontentsline{toc}{section}{Acknowledgment}
 The author is thankful to Prof. D. Mukhopadhaya of Electronics
 and Tele-Communication Engineering Department, Jadavpur University, India for
 fruitful discussion and suggestions on the draft manuscript
 of the paper.

 \bibliographystyle{IEEEtran}

\end{document}